\def\year{2021}\relax

\documentclass[letterpaper]{article} 
\usepackage{aaai20}  
\usepackage{times}  
\usepackage{helvet}  
\usepackage{courier}  
\usepackage{url}  
\usepackage{graphicx}  
\usepackage{subfigure}
\usepackage{enumerate}
\usepackage{hyperref}
\usepackage{array}
\usepackage{dblfloatfix} 
\usepackage{graphics}

\begin{document}
%
\title{MeetSum: Transforming Meeting Transcript Summarization using Transformers!}
\author{Nima Sadri, Bihan Liu, Bohan Zhang\\
\{nsadri, b263liu, b327zhan\}@uwaterloo.ca\\
University of Waterloo\\
Waterloo, ON, Canada\\
}
\maketitle



\begin{abstract}

Creating abstractive summaries from meeting transcripts has proven to be challenging due to the limited amount of labeled data available for training neural network models. Moreover, Transformer-based architectures have proven to beat state-of-the-art models in summarizing \textit{news} data. In this paper, we utilize a Transformer-based Pointer Generator Network to generate abstract summaries for meeting transcripts. This model uses 2 LSTMs as an encoder and a decoder, a Pointer network which copies words from the inputted text, and a Generator network to produce out-of-vocabulary words (hence making the summary abstractive). Moreover, a coverage mechanism is used to avoid repetition of words in the generated summary. First, we show that training the model on a news summary dataset and using zero-shot learning to test it on the meeting dataset proves to produce better results than training it on the AMI meeting dataset. Second, we show that training this model first on out-of-domain data, such as the CNN-Dailymail dataset, followed by a fine-tuning stage on the AMI meeting dataset is able to improve the performance of the model significantly. We test our model on a testing set from the AMI dataset and report the ROUGE-2 score of the generated summary to compare with previous literature. We also report the Factual score of our summaries since it is a better benchmark for abstractive summaries since the ROUGE-2 score is limited to measuring word-overlaps. We show that our improved model is able to improve on previous models by at least 5 ROUGE-2 scores, which is a substantial improvement. Also, a qualitative analysis of the summaries generated by our model shows that these summaries and human-readable and indeed capture most of the important information from the transcripts.




\end{abstract}


\section{Introduction}
$\;$
With the wide adoption of remote-work in 2020 and 2021, more than ever meetings are being conducted virtually. Zoom and Google Hangouts reported 300 million and 100 million daily meetings conducted in 2020, respectively. Most major video conferencing software, such as Zoom, Teams, Google Hangouts, and Skype auto-generate transcripts of the meeting using text-to-speech technology. Currently these transcripts are mostly being used as live sub-titles. Some teams, however, keep the transcripts and use it to refer back to as notes. These transcripts are usually very long and reading through them can be a lengthy process. However, meeting transcripts can be used to generate a short summary of the meeting that all the employees and managers can refer back to. \\

Summarization can happen using either an extractive or abstractive approach. In extractive summarization, most important phrases from the inputted text are chosen and stitched together to form a summary. In the abstractive approach, however, we are able to add words that are not directly in the inputted text but imply the same idea. Creating a useful, condensed, and readable summary usually requires abstractive summarization techniques and lots of training data.  \\

Due to the confidential nature of business meetings, available meeting transcripts and their summaries are very limited, with largest training datasets consisting of only a few hundred data points. This often leads to models that do not perform well. However, there are other datasets from news articles and their summaries consisting of hundreds of thousands data points. We will try to improve the performance of current models attempting to summarize meeting notes by training our model using both news article data and meeting transcript data. We will then compare this using a model trained only with meeting transcript data to see if our modifications improved the model. Both models use the same algorithm, a Pointer-Generator method equipped with a coverage mechanism \cite{DBLP:journals/corr/SeeLM17}. The Pointer-Generator method is able to both copy words from the input text and add new words from a fixed vocabulary. The mechanism responsible for adding new words is a sequence-to-sequence model using attention, while the mechanism responsible for copying words from the input text is inspired by a pointer network. The sequence-to-sequence mechanism encodes the inputted text first. It then uses the generated encoding, along with a learned attention distribution and a decoder to create the summary. The attention distribution allows the model to find relationships between words in the given text and summarize them using a word that is not included in the inputted text. The aforementioned coverage mechanism prohibits the model from unnecessarily repeating the same phrase over and over again. \\

Our model will generate multi-sentence summaries. The meeting transcript dataset used for training and testing is the AMI Meeting Corpus dataset.  We will also use summaries generated from CNN-Dailymail articles to train our model. We will test our models on meeting transcript notes, by measuring the Factual score of the outputted summaries \cite{DBLP:journals/corr/abs-1910-12840} and the ROUGE-2 $F_1$ score. \\

\noindent{{\bf Main Results}}

After the experiment, we found that the combination of news data and meeting data makes an overall way better meeting transcript summarization than simply using news data or meeting data. The generated text is human-readable and concise while keeping all the key points of the meeting.\\

We were able to show that even with very limited resources of meeting transcript data, it is reasonable to combine other datasets to generate better summaries.\\

\noindent{{\bf Contributions}}

\noindent Our contributions include:
\begin{itemize}
    \item Modifying the implementation of an existing model architecture in order to save the actual summaries for later inspection.
    \item Implementing an algorithm to process any \texttt{.csv} data to match the inputs of our models.
    \item For the first time, to our knowledge, used the Factual score to evaluate the performance of our abstract summarizations on the AMI dataset
    \item Showed that a simple model trained with out-of-domain data can be easily fine-tuned to generate high quality meeting transcripts
    \item Found a positive correlation between the accuracy of the summaries and the closeness of data-set and actual text relationship.
\end{itemize}


\section{Related Work} 

\subsection{Abstractive Summarization}
$\;$
A significant amount of work has been done on automatic text summarization. Considering the fact that the earlier models mostly concentrated on extractive methods like the BERTSUM pre-trained model \cite{DBLP:journals/corr/abs-1903-10318} which contains several summarization layers that can be applied with BERT, abstractive summarization is a comparatively novel research field with less progress. Nevertheless, some recent advancements have been made using attention or attention-based deep recurrent neural networks as methods to generate abstractive summaries, achieving state-of-art results.\\

Some previous models used the Transformer architecture, which is solely based on the attention mechanism \cite{DBLP:journals/corr/VaswaniSPUJGKP17}. By only relying on the attention mechanism, the model is able to remove the recurrence and convolutions entirely from the implementation. Hence, these models are proven to be superior in quality while being more parallelizable and requiring significantly less time to train than traditional deep neural networks \cite{Yu2017SummarizationWA}. \\

In 2015, the attention-encoder mechanism was first used for abstractive text summarization \cite{DBLP:journals/corr/RushCW15}, achieving then state-of-art performance on two-sentence summarization. Later in 2016, encoder-decoder RNNs with attention mechanism \cite{DBLP:journals/corr/NallapatiXZ16} outperformed the previous models on abstractive summarization for the same dataset.

\subsection{Meeting Transcript Summarization}
$\;$
To synthesize abstractive summaries from meeting transcripts, a model was suggested that combines important content from several utterances, separating the meeting transcripts into various topic segments and then identifies the important utterances in each segment using a supervised learning approach \cite{DBLP:journals/corr/BanerjeeMS16}. This model significantly improves the readability of the summary without losing any important information. For example, if a sentence contains information about plants and computers, the model separates the phrases about plants from those about computers, and then extracts key information from each topic. A directed graph is created from merging each utterances' dependency parses, and then Integer linear programming (ILP) obtains the most informative and well-formed summary. This formulation reduces disfluencies by leveraging grammatical relations that are more prominent in the non-conversational style of text, comparable to human-written abstractive summaries. In addition, based on the previous 2015 model \cite{DBLP:journals/corr/RushCW15}, a novel model proposed improvements by utilizing a hybrid pointer-generator network that can copy words from the source text via pointing and a coverage mechanism to keep track of the words used in the summary to avoid repetition \cite{DBLP:journals/corr/SeeLM17}. The architecture reinforces the accurate reproduction of information and discourages repetition.\\

Enforcing a constraint on the length of an abstractive summary generated by a model is difficult, particularly when training data is scarce. At times, it may be helpful to be able to interpret how a model generates a summary, particularly for our task. Other authors suggest a Multi-level Summarizer (MLS) that can achieve a controllable-length abstractive summary in a way that can be interpreted by humans easily \cite{sarkhel2020interpretable}. MLS uses a multi-headed attention mechanism, similar to other models\cite{DBLP:journals/corr/abs-2002-07845}, alongside independent kernels, each of which can be synthetically or semantically interpreted. MLS performs better than other baseline models using quantitative metrics. \\

The models implemented and the corresponding datasets used in this paper are inspired by the six papers mentioned above.


\section{Methodology}

\subsection{Algorithm}
$\;$
The algorithm we used is a pointer-generator algorithm suggested by \cite{DBLP:journals/corr/SeeLM17}. An initial sequence-to-sequence model using attention is used and then modified by adding a pointer-generator network and a coverage mechanism. These are explained below. \\

We are given a multi sentence text that we need to summarize. Each word is embedded as a vector using a positional embedding. The $i_{th}$ word is embeded as $w_i$. We start by explaining the core of the algorithm first and then introduce a pointer-generator network and the coverage method to improve the model \cite{DBLP:journals/corr/SeeLM17}. \\

$\;$

\noindent \textbf{3.1.1 Core Model}

$\;$

Encoder is a one-layered bidirectional LSTM. Each $w_i$ is given to the encoder in order that it appears in the input text. Encoder uses $w_i$ to create a hidden state $h_i$. \\

Decoder is a unidirectional LSTM. At time $t$, decoder receives the previous word either from the reference summary (during training) or by the previously predicted word (during testing). We use $s_t$ to denote the state of the decoder at time $t$. \\

\noindent Decoder calculates the attention at time $t$ as $$a_i^t = \textrm{softmax}(e_i^t = v^\top tanh(W_h h_i + W_s s_t + b_{attn}))$$ The attention is basically a probability distribution among the words in the input document, which the decoder uses to see which words to pay the most "attention" to while predicting the next word in the summary.
The context vector is calculated as $h_i^* = \sum_{i}{a_i^t h_i}$. Vocabulary probability distribution is then calculated as 
$$ P_{\textrm{vocab}} = \textrm{softmax}(V^\prime(V [s_t h_t^* ]+ b) + b^\prime))$$ where $V$, $V^\prime$, $b$, $b^\prime $ can be learned. The prediction for the next word is then picked from the probability distribution $P_\textrm{vocab}(w)$. \\

We note that the loss function used in the training (at time $t$) is as follows:

$$ loss_t = -log P(w_t^*) $$

The total loss is the average of losses. That is,

$$ loss = \sum_{t=1}^{T}{loss_t} $$

$\;$

\noindent \textbf{3.1.2 Adding Pointer-Generator}

$\;$

At step $t$, generation probability $p_{gen}$ is calculated as follows:

$$p_{gen} = \sigma (w_{h^*}^\top h^*_t + w_s^\top s_t + w_x^\top x_t + b_{pointer})$$ \\

\noindent Recall that
\begin{itemize}
\item $h_t^*$ is context vector calculated in previous step
\item $s_t$ is the decoder state
\item $x_t$ is the input to the decoder 
\item $w_{h^*}$, $w_s$, $w_x$ are vectors that the network will learn
\item $b_{pointer}$ is a real number that can be learned
\item $\sigma(x) = \frac{1}{1+e^{-x}}$
\end{itemize}

$\;$

The model chooses to generate a new word with probability $p_{gen}$ and chooses to copy a word from the input text with probability $1-p_{gen}$. To copy a word, the model uses the attention distribution $a_t$ computed previously. To generate a new word, the model samples a new word from the  $P_{\textrm{vocab}}$ distribution.\\

The addition of the pointer-generator network, therefore, allows the model in addition to copying a word from the input text to also generate out-of-vocabulary words and include them in the summary.

$\;$

\noindent \textbf{3.1.3 Adding Coverage}

$\;$

When generating multi-sentence summaries, most models usually tend to keep repeating a word or a group of words. We will use the coverage mechanism suggested by a previous paper \cite{tu2016modeling}. to fix this issue. Let $c_0 = 0$ and $c_t = c_{t-1} + a^{t-1}$ be the coverage vector (at time $t$). Equivalently, $c_t = \sum_{time=0}^{t-1}{a^{time}}$. This vector encompasses how much attention our model has given to words in the input text while generating our summary output until this point. We will use the coverage vector when calculating attention at each step $t$. Therefore, the formula for attention shall be modified as follows
$$
a_i^t = \textrm{softmax}(e_i^t = v^\top tanh(W_h h_i + W_s s_t + w_c c_i^t+ b_{attn}))
$$
Note that $w_c$ is learned along with previously mentioned parameters. Hence when the attention mechanism wants to choose the next word from the input sentence to attend to, it will take into account the previously attended words in previous time-steps. To penalize the network for attending to the same word over and over again, we shall define a coverage loss, which will be integrated into our original loss function. 
$$
\textrm{coverageLoss}_t = \sum_i{min(a_i^t, c_i^t)}
$$
Since $\sum_i{a_i^t} = 1$, we can conclude that $\textrm{coverageLoss}_t \leq 1$ 

We then use the following loss function: 
$$
\textrm{loss}_t = -log P(w_t^*) + \textrm{coverageLoss}_t
$$

$\;$
\subsection{Data Mining}
$\;$

Since we want to measure the performance of training the algorithm with different datasets, we picked two datasets to train and test our data.

$\;$

\subsection{datasets}

$\;$

The following datasets are used in our experiment: \\

\noindent \textbf{3.3.1 AMI Meeting Corpus} \cite{Mccowan05theami}

$\;$

This dataset contains 142 transcripts of meetings, each with 4 to 5 attendants, and their abstract summaries. Each transcript is around 10,000 words long, and each summary is around 100 words long.

We divide the AMI dataset into 3 disjoint parts:
\begin{itemize}
\item A training set consisting of 70\% of the data (\texttt{AMI\_train})
\item A validation set consisting of 15\% of the data (\texttt{AMI\_validation})
\item A test set consisting of  the remaining 15\% of the data (\texttt{AMI\_test})
\end{itemize}

$\;$

\noindent \textbf{3.3.2 CNN-DailyMail} \href{https://github.com/abisee/cnn-dailymail}{\cite{DBLP:journals/corr/SeeLM17}} 

$\;$

This dataset contains over 300,000 news articles from CNN and Dailymail. Each article also contains a summary written by a human journalist. Each news article text is around 750 words long, and its multiple sentences summary is around 50 words long. 

We shall refer to this dataset as \texttt{CNN-Dailymail}, and we note that all this data will only be used for training our model and not for testing it.

$\;$









$\;$
\subsection{Why These Datasets?}

$\;$

The algorithm we use has already been used to successfully summarize text in English. Thus, we choose these datasets, which are written in English, to assure the algorithm runs successfully. Also, the summarization in these three datasets are all human-written, which makes the dataset more reliable to train a model which would generate human-readable summarizations. Besides that, both datasets have abstract summarizations, which is consistent with the algorithm we use.

In particular, we believe the news contains millions of words involving different scenarios, and hence it may help the model learn to adjust the word embedding space in order to perform better on meeting transcript. 

Also, note that when generating a news summarizations, the model should generate a short summary without losing the key information. Since this property is rather important in summarizing meeting transcripts, we believe that the \texttt{CNN-Dailymail} dataset can teach our model to generate summaries that captures all the key information from the meeting. 

$\;$

\subsection{Experiment Process}

$\;$

We first separate the AMI dataset to \texttt{AMI\_train}, \texttt{AMI\_validation}, \texttt{AMI\_test} following a ratio of 70:15:15, as suggested previously. \\

We will train 3 models, all of which share the same architecture and hyperparameters. The only difference between these models are the data that is used to train them.

For our first baseline model, we will use the algorithm explained above and train it with \texttt{AMI\_train}. This model shall be referred to as the \textsc{meetingOnly} model since it is only trained with the \texttt{AMI} meeting dataset. \\

Our second baseline model is only trained on the \texttt{CNN-Dailymail} dataset. This model shall be referred to as the \textsc{newsOnly} model since it is only trained with the \texttt{CNN-Dailymail} news dataset. \\

For our improved model, we will fuse the same algorithm but we will train it on the \texttt{CNN-Dailymain} dataset first and then fine-tune it by training it on the \texttt{AMI\_train} dataset. This model will be referred to as the \textsc{advancedModel}. \\

The authors of the Pointer-Generator network model found that using a learning rate of 0.15 works best \cite{DBLP:journals/corr/SeeLM17} when training on the \texttt{CNN-Daily} dataset. Hence, we will also use the same learning rate.



$\;$


$\;$

\subsection{Evaluating the performance}

$\;$

After the experiment, we will run our model on \texttt{AMI\_test}. Since the generated text is an abstract summarization, the generated text might use completely different words and grammar to express the same meaning as the target text. If we use the ROUGE score, which is a measurement of performance based on the same words between two texts, then the result could be poor even two sentences are expressing the same meaning. Thus, we will mainly use the Factual score \cite{DBLP:journals/corr/abs-1910-12840} instead which can better capture the abstract meaning between two pieces of text. We will report the ROUGE-2 score as well to be able to compare our models with previous benchmarks.

$\;$

\noindent The Factual score is calculated as follows:
\begin{enumerate}
    \item \texttt{OpenIE} consume the text and generates pairs of facts in the form (argument, predicate, argument)
    \item Google universal sentence encoder consumes the fact pairs and encode them in the embedding space.
    \item Score can be evaluated based on the cosine-similarity of each pair of vector $G_i$, $R_j$, in the fact embedding space. The similarity is $s = \frac{G_i * R_j}{||G_i||*||R_j||}$.
\end{enumerate}

$\;$

In particular:

$$ Precision = Fact\_P = \frac{1}{m}\sum_{i=1}^m{max_{j=1}^{n}s_{ij}} $$

Where 
$$ Recall = Fact\_R = \frac{1}{n}\sum_{i=1}^n{max_{j=1}^{m}s_{ij}} $$

$$ F1 = Fact\_F1 = \frac{2*Fact\_P * Fact\_R}{Fact\_P + Fact\_R} $$ \\

Similar to the confusion matrix, also known as the error matrix, the $F1$ score is seeking a balance between Precision and Recall, so we can adjust our model based on $F1$ score. And finally use these three values as a measurement of performance of our model.

\section{Result}

\subsection{Implementation}
$\;$
We shall measure the ROUGE-2 $F_1$ score using the \texttt{rouge}\footnote{\href{https://github.com/pltrdy/rouge}{\texttt{https://github.com/pltrdy/rouge}}} library. We also updated the official implementation of the Factual score \footnote{\href{https://github.com/yuhui-zh15/nlg\_metrics}{\texttt{https://github.com/yuhui-zh15/nlg\_metrics}}} to work with newer versions of \texttt{OpenIE}. \\
$\;$

We use a fork\footnote{\href{https://github.com/becxer/pointer-generator/}{\texttt{https://github.com/becxer/pointer-generator/}}} of the official implementation of the Pointer-Generator network \cite{DBLP:journals/corr/SeeLM17}. The original codebase \footnote{\href{https://github.com/abisee/pointer-generator}{\texttt{https://github.com/abisee/pointer-generator}}} is written in \texttt{python2}, but the fork we used was updated to use \texttt{python3}. Both codebases have been written four years ago, and as a result both were using version $1$ of Tensorflow. We updated the code to use Tensorflow $2$. Our code is publicly available on GitHub
\footnote{\href{https://github.com/nimasadri11/CS480}{\texttt{https://github.com/nimasadri11/CS480}}}. \\

$\;$

The \textsc{newsOnly} model was trained by authors of the Pointer Generator network \cite{DBLP:journals/corr/SeeLM17} on the \texttt{CNN-Dailymail} dataset and can be downloaded from their GitHub repository. We use this already trained model for our baseline. The model was trained using Adagrad with learning rate $\eta = 0.15$. This combination was found to work the best \cite{DBLP:journals/corr/SeeLM17} compared other candidates such as Adam, RMSProp, and Stochastic Gradient Descent. The model was trained for 13 epochs (or 238,000 steps) with a batch size of 16. The coverage mechanism was added during the last 3,000 steps of training with $\lambda = 1$. Other values for $\lambda$ were tried, but ultimately $\lambda = 1$ proved to work the best \cite{DBLP:journals/corr/SeeLM17}. We trained the \textsc{meetingOnly} model with the same hyper-parameters for about 5,000 steps, after which the loss stopped decreasing. Our fine-tuned model (\textsc{advancedModel}) was trained with the same hyper-parameters for about 1,000 steps. Training more did not decrease the validation loss (on \texttt{AMI\_validation}), so we used the checkpoint for the model after 1,000 training steps. This fine-tuning training process took about 2 hours on a single Nvidia T4 GPU. \\

$\;$

\begin{figure}[htbp!]
  \centering
  \includegraphics[width=0.9\linewidth]{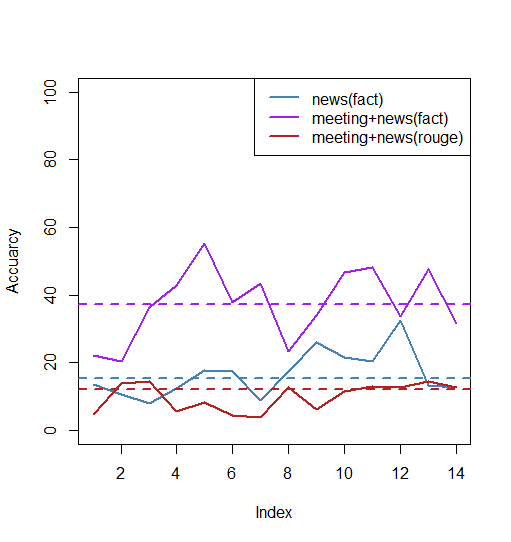}
  \caption{The Summary Score Comparison Plot}
  \label{fig:results3}
\end{figure}

\begin{figure}[htbp!]
  \centering
  \includegraphics[width=0.9\linewidth]{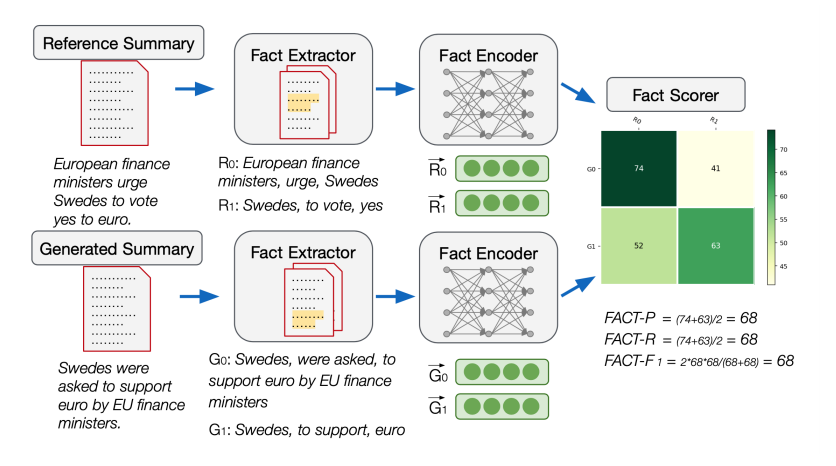}
  \caption{Computation of the Factual Score.}
  \label{fig:results2}
\end{figure}
\begin{figure*}[b!]
    \centering
  \includegraphics[width=0.95\linewidth]{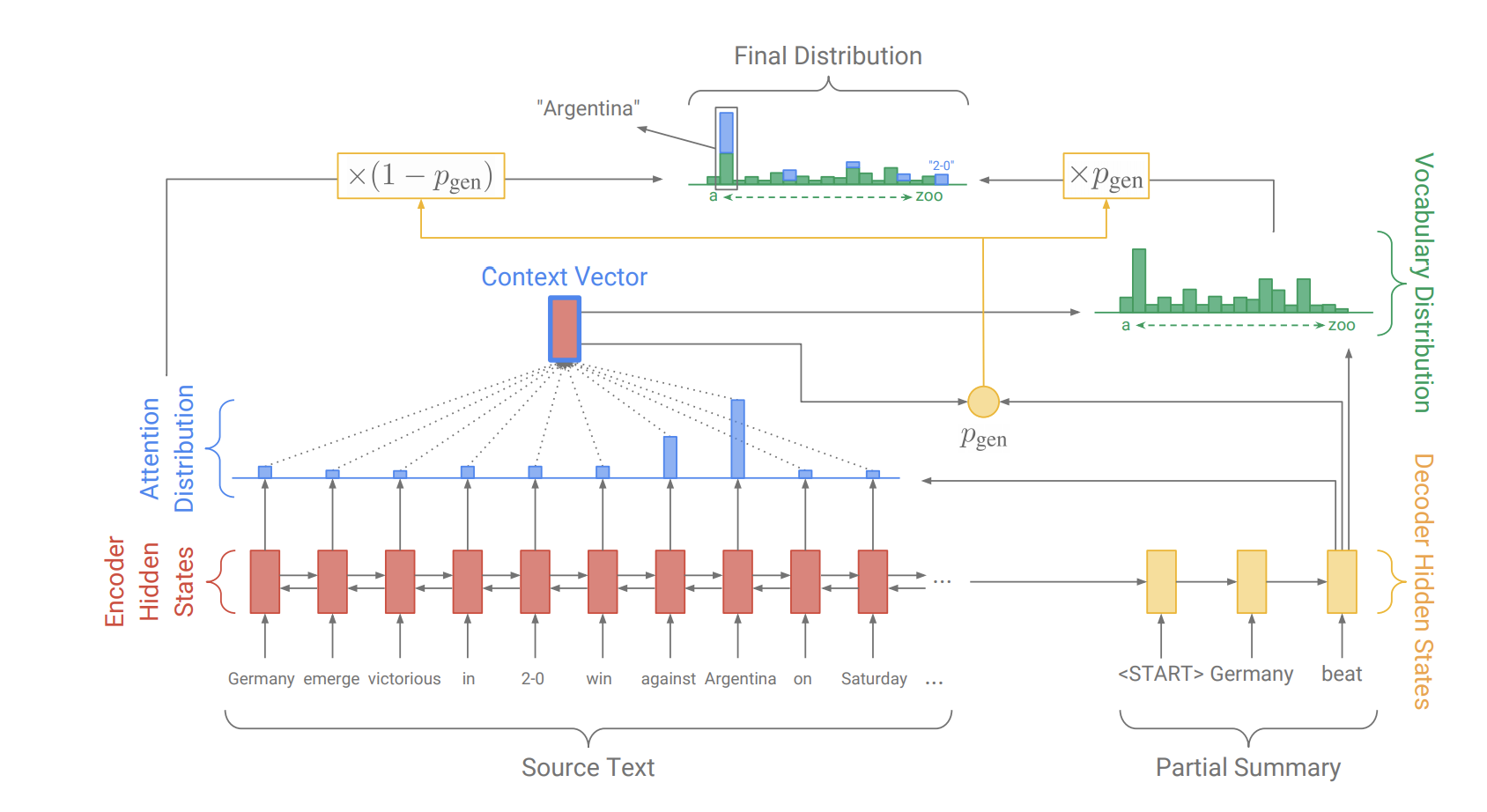}
  \caption{The Pointer-Generator Networks}
\end{figure*}
\subsection{Performance}
$\;$
We use two measurements to evaluate summaries generated by our models. The Factual score will reflect how well the summary captures the abstractive meaning of the original text. ROUGE-2 score is also reported since it is a popular measurement and is required to compare the performance of our algorithm with other models. The Factual scores for all models are reported in Table 1. The ROUGE-2 $F_1$ scores are reported in Table 2. A plot of score of the each measurement for 14 generated summaries with their median in dashed lines are visualized in Figure 3. Our \textsc{advancedModel} surpassed the \textsc{newsOnly} model in both the ROUGE-2 score and the Factual Score. However, the model that was only trained on the \texttt{AMI} meeting dataset performed terrible, reaching a Factual score of $0$. This issue is discussed in Section $5.1$.

\begin{table}[h!]
\centering
\resizebox{\columnwidth}{!}{
\begin{tabular}{|c|c|c|c|}

\hline
Fact Score & \textsc{meetingOnly} & \textsc{newsOnly} & \textsc{advancedModel} \\
\hline
Min & \~0 & 7.87 & 20.48 \\
Median & \~0 & 15.49  & 37.80\\
Mean & \~0 & 16.59  & 37.97\\
Max & \~0 & 32.64 & 55.21\\
\hline

\end{tabular}
}
\caption{Factual Scores}
\end{table}

As seen in Table 1, all 4 statistic values of the factual scores of our \textsc{advancedModel} trained with the combination of meeting and news data are significantly greater than the values for the other two models that were trained on only one of the datasets. This can also be seen in Figure 1.

\begin{table}[h!]
\centering
\resizebox{\columnwidth}{!}{
\begin{tabular}{|c|c|c|c|}
\hline
ROUGE-2 & Our model & Mehdad et al. & Banerjee et al. \\
\hline
Min & 3.91 & - &-\\
Median & 12.03&- &- \\
Mean & 9.91 & 4.0-4.8 & 3.6-4.8\\
Max & 14.61&- &-\\
\hline
\end{tabular}
}
\caption{ROUGE-2 Scores}
\end{table}

According to the results provided in Figure 1 and Table 1, we conclude that the combination of meeting data and news data significantly increased the Factual score. \\

An example of a generated summary by our \textsc{advancedModel} is shown in Table 3 (Factual score $= 40$).
\begin{center}
\begin{table}[h!p]
{\def\arraystretch{1.3}
\begin{tabular}{|>{\centering\arraybackslash}m{2.2cm}||>{\centering\arraybackslash}m{5.2cm}|}
\hline
 Model & Summaries \\
\hline
\textsc{newsOnly} & points will be assigned to gather the data from programme or contents.
we can think about an interface with uh well yeah.
"we can think about an interface with well yeah." says norman.  \\
\hline
\texttt{advanced\ Model} &the industrial designer gave a presentation on the working design with the group and discussed their preference agenda.
the industrial designer gave his presentation on the basic components of the remote , and the interface designer, presented two existing remote components.
the user interface designer discussed the interior workings of the remote and the industrial designer, the industrial designer, and the industrial designer presented.
the project manager discussed the projected price point and price point.
the marketing expert presented an evaluation of the prototype and the group discussed their experiences with each group.\\
\hline
Human-written & the project manager introduced the project to the group.
the group set an agenda for the meeting and discussed the materials sent to them by the account manager.
they discussed and explained their roles in the project.
the group began a discussion about their initial ideas for the product.
they discussed several usability features: adding speech recognition and an option to choose what to watch by channel or by content, reducing the number of buttons by using the television screen to display options, and adding a light adaptation system.
all participants were instructed to gather more information for the next meeting, the functional design meeting.\\
\hline
\end{tabular}
}
\caption{Sample Summary Results}
\end{table}
\end{center}

The summary generated by \textsc{newsOnly} model does not seem to capture the main points of the meeting. However, the summary generated by \textsc{advancedModel} does indeed capture most of the main points of the meeting, and this summary is relatively comparable or even more concise than the one written manually by a human.

\subsection{Lessons Learned}
$\;$
We learned that there are difficulties when using outdated code that use old dependencies.  We also learned that pretraining neural networks models with out-of-domain data is extremely advantageous in cases where a limited amount of domain data is available.
Moreover, we got familiar with the variety of datasets in text summarization. We also learned how to efficiently pre-process our data from different datasets in a form that fits the schema of our model. \\


\section{Discussion}



\subsection{Interpretation}
$\;$
As we hypothesized, fine-tuning the Pointer-Generator network \cite{DBLP:journals/corr/SeeLM17} with even a small amount of meeting summary data, significantly improved the performance of our model when evaluated on the abstractive meeting summarization task. In fact, the summary generated by the baseline model (\textsc{newsOnly} model) only captured $\sim$16\% of the important information when evaluated with the Factual score. However, the same model when fine-tuned with $\sim$100 meeting transcript summary data points (i.e. the \textsc{advancedModel}) was able to capture more than double ($\sim$37\%) of the important information from the original text when evaluated with the same metric. Some summaries generated by the \textsc{newsOnly} model were shorter compared to the ones generated by the \textsc{advancedModel} (e.g. the summary in Table 3). However, the length of the summaries generated by both models in most cases were about the same. 

$\;$

Although we expected to see an improvement in the fine-tuned model, we did not expect the model to outperform the base model to this degree. This was due to the fact that the fine-tunning process only included $\sim$100 new data points, compared to the $\sim$300,000 news data points that both of the models were pre-trained on. This shows the significance of fine-tunning a pre-trained model (even with a small amount of data).

\subsection{Implications}
$\;$
Similar to the state-of-the-art results the Transformer mechanism \cite{DBLP:journals/corr/VaswaniSPUJGKP17} has been able to achieve in a wide variety of natural language processing tasks (e.g. machine translation, summarizing news data, etc.). In this paper, we saw that the Transformer mechanism, when adapted to be used in an abstractive summarization task using the Pointer Generator network \cite{DBLP:journals/corr/SeeLM17} is able to outperform other models in generating abstract summaries for the \texttt{AMI} meeting dataset when compared using the ROUGE-2 score. These other models were not evaluated using the Factual score; hence we are only able to compare them with respect to the ROUGE score. Compared to the entailment/fusion approach \cite{mehdad-etal-2013-abstractive} or the graph dependency model \cite{DBLP:journals/corr/BanerjeeMS16}, our model improves on the ROUGE-2 F-Score by 5 points, which is a substantial improvement.

$\;$

Moreover, we see that the Pointer Generator network \cite{DBLP:journals/corr/SeeLM17} reaches comparable Factual scores when:
\begin{itemize}
\item trained on news dataset and evaluated on the same new dataset (Factual score of $\sim 43$) \cite{DBLP:journals/corr/abs-1910-12840}
\item pre-trained on news dataset, fine-tuned on meeting dataset (with $\sim$100 data points), and evaluated on the meeting dataset
\end{itemize}
$\;$

This is despite that the Pointer Generator network has been fine-tuned on a very small number ($\sim 100$) of data points. Therefore, we can conclude that the pre-trained Pointer Generator network is able to perform comparably well on out-of-domain datasets, if it is fine-tuned with a very small amount of data.

$\;$

Contrary to the state-of-the-art results that the Pointer Generator network \cite{DBLP:journals/corr/SeeLM17} achieves when pre-trained on the news data, this network is incapable of learning with small amounts of data. This was evident when, instead of using the model pre-trained on the news data, we trained the model {\bf only} on meeting data. The summaries generated by this model (i.e. the \textsc{meetingOnly} model) consisted of copying 1-2 words from the input and repeating it numerous times. Even when increasing the weight of the coverage mechanism to penalize the model more for repeating the same words, the problem persisted. This, we believe, is due to the small amount of data this model was trained on ($\sim$ 100 data points). 

$\;$

We also identify the vast practical implication that our model can have in today's online workplace. 
Since most meetings happen online in today's remote workplace, and most popular video conferencing software already generated automated transcripts, an automated summary of the meetings can be generated with our model, which can then be sent to all meeting attendees and other managers. We acknowledge that this has been possible before; however, we recognize that our model allows for substantially better automated summaries, and the summaries generated by our model contains important information from the meeting to a higher degree when compared to previously established models.

\subsection{Limitations}
$\;$
We note that our model has only been evaluated on summarizing the meeting transcripts from the \texttt{AMI} corpus. This dataset only contains transcripts from product planning meetings. We were unable to evaluate our model on other types of meetings, since we were not able to find datasets that includes meeting transcripts (and summaries) from other types of meetings. Evaluating this model with other meeting transcripts is an interesting direction for our future work if we are able to find other meeting transcript datasets.

$\;$

Also, we did not evaluate our model on non-meeting datasets such as the news dataset, after training it with the meeting dataset. We speculate that our model's performance to generate summaries for news data has not been reduced since we only fine-tuned the model with $\sim$ 100 non-news data points. However, we have not verified this hypothesis. This can also be a future work for us to investigate. 

$\;$

Finally, we recognize that our test set included only 15 data points. As such, these results may not generalize well. However, this is a limitation in most of the work in meeting transcript summarization work in the literature since no large dataset in this domain exists to our knowledge.

\section{Conclusion}
$\;$
The number of online meetings has increased significantly on video conferencing applications such as Zoom and Google Meet. Most of these applications generate an automatic transcript of the meeting which can be download. However, these transcripts are usually very long and summarizing them for future reference is a valuable yet tedious task. Currently the largest datasets of meeting transcript and their summaries consists of only a hundred data points, which is not enough data for most neural network models to learn sufficiently well. In this paper, we explored how training a Transformer-based neural network model (i.e. the Pointer Generator network) first with out-of-domain (e.g. news) datasets and then fine-tuning it with meeting transcript summaries is able to outperform other models (\cite{mehdad-etal-2013-abstractive} and \cite{DBLP:journals/corr/BanerjeeMS16}) by 5 ROUGE-2 scores. Moreover, we showed that the out-of-domain pre-training has a substaintial impact on the generated summaries. This is evident by reading the summaries generated by the models and also the 20 point increase in the Factual score when the out-of-domain pre-training occurs.\\

$\;$

We outline to interesting future-directions in this area below:
\begin{itemize}
\item Measuring the performance of other models when evaluated on the Factual. We hope that future researchers adopt the Factual score when evaluating their models on a abstractive summarization task, since it is better able to capture abstract meanings of the texts than the currently widely adopted ROUGE score, which only captures the extractive similarities. 
\item Using different learning rates during the pre-training and fine-tuning phase. The fine-tuning phase of our task included much less training data but the fine-tuned dataset is much more similar to the testing dataset than the pre-training dataset (i.e. the news summaries). Therefore, it would be interesting to measure the effect of using a larger learning rate during the fine-tuning process.
\item Publicly releasing a larger dataset of meeting transcripts and human written summaries can vastly improve the performance of most models. We encourage researchers that have access to such datasets to make them publicly available (of course after acquiring ethical copyrights)
\item We suspect that further improvement can also be gained by replacing the LSTM encoder with a BERT encoder. This is a particularly interesting future direction that has been explored for news summarization, but, to our knowledge, it has not yet been tested for meeting transcript summarization.
\end{itemize}

We are optimistic about the future of text summarization using Transformer-based models and hope that more interesting work is done in the meeting transcript summarization space. 

\newpage
\bibliographystyle{aaai}
\bibliography{report}

\begin{thebibliography}{}

\bibitem[\protect\citeauthoryear{Banerjee, Mitra, and
  Sugiyama}{2016}]{DBLP:journals/corr/BanerjeeMS16}
Banerjee, S.; Mitra, P.; and Sugiyama, K.
\newblock 2016.
\newblock Generating abstractive summaries from meeting transcripts.
\newblock {\em CoRR} abs/1609.07033.

\bibitem[\protect\citeauthoryear{Kryscinski \bgroup et al\mbox.\egroup
  }{2019}]{DBLP:journals/corr/abs-1910-12840}
Kryscinski, W.; McCann, B.; Xiong, C.; and Socher, R.
\newblock 2019.
\newblock Evaluating the factual consistency of abstractive text summarization.
\newblock {\em CoRR} abs/1910.12840.

\bibitem[\protect\citeauthoryear{Liu}{2019}]{DBLP:journals/corr/abs-1903-10318}
Liu, Y.
\newblock 2019.
\newblock Fine-tune {BERT} for extractive summarization.
\newblock {\em CoRR} abs/1903.10318.

\bibitem[\protect\citeauthoryear{Mccowan \bgroup et al\mbox.\egroup
  }{2005}]{Mccowan05theami}
Mccowan, I.; Lathoud, G.; Lincoln, M.; Lisowska, A.; Post, W.; Reidsma, D.; and
  Wellner, P.
\newblock 2005.
\newblock The ami meeting corpus.
\newblock In {\em In: Proceedings Measuring Behavior 2005, 5th International
  Conference on Methods and Techniques in Behavioral Research. L.P.J.J. Noldus,
  F. Grieco, L.W.S. Loijens and P.H. Zimmerman (Eds.), Wageningen: Noldus
  Information Technology}.

\bibitem[\protect\citeauthoryear{Mehdad \bgroup et al\mbox.\egroup
  }{2013}]{mehdad-etal-2013-abstractive}
Mehdad, Y.; Carenini, G.; Tompa, F.; and Ng, R.~T.
\newblock 2013.
\newblock Abstractive meeting summarization with entailment and fusion.
\newblock In {\em Proceedings of the 14th {E}uropean Workshop on Natural
  Language Generation},  136--146.
\newblock Sofia, Bulgaria: Association for Computational Linguistics.

\bibitem[\protect\citeauthoryear{Nallapati, Xiang, and
  Zhou}{2016}]{DBLP:journals/corr/NallapatiXZ16}
Nallapati, R.; Xiang, B.; and Zhou, B.
\newblock 2016.
\newblock Sequence-to-sequence rnns for text summarization.
\newblock {\em CoRR} abs/1602.06023.

\bibitem[\protect\citeauthoryear{Rush, Chopra, and
  Weston}{2015}]{DBLP:journals/corr/RushCW15}
Rush, A.~M.; Chopra, S.; and Weston, J.
\newblock 2015.
\newblock A neural attention model for abstractive sentence summarization.
\newblock {\em CoRR} abs/1509.00685.

\bibitem[\protect\citeauthoryear{Sarkhel \bgroup et al\mbox.\egroup
  }{2020a}]{sarkhel2020interpretable}
Sarkhel, R.; Keymanesh, M.; Nandi, A.; and Parthasarathy, S.
\newblock 2020a.
\newblock Interpretable multi-headed attention for abstractive summarization at
  controllable lengths.

\bibitem[\protect\citeauthoryear{Sarkhel \bgroup et al\mbox.\egroup
  }{2020b}]{DBLP:journals/corr/abs-2002-07845}
Sarkhel, R.; Keymanesh, M.; Nandi, A.; and Parthasarathy, S.
\newblock 2020b.
\newblock Transfer learning for abstractive summarization at controllable
  budgets.
\newblock {\em CoRR} abs/2002.07845.

\bibitem[\protect\citeauthoryear{See, Liu, and
  Manning}{2017}]{DBLP:journals/corr/SeeLM17}
See, A.; Liu, P.~J.; and Manning, C.~D.
\newblock 2017.
\newblock Get to the point: Summarization with pointer-generator networks.
\newblock {\em CoRR} abs/1704.04368.

\bibitem[\protect\citeauthoryear{Tu \bgroup et al\mbox.\egroup
  }{2016}]{tu2016modeling}
Tu, Z.; Lu, Z.; Liu, Y.; Liu, X.; and Li, H.
\newblock 2016.
\newblock Modeling coverage for neural machine translation.

\bibitem[\protect\citeauthoryear{Vaswani \bgroup et al\mbox.\egroup
  }{2017}]{DBLP:journals/corr/VaswaniSPUJGKP17}
Vaswani, A.; Shazeer, N.; Parmar, N.; Uszkoreit, J.; Jones, L.; Gomez, A.~N.;
  Kaiser, L.; and Polosukhin, I.
\newblock 2017.
\newblock Attention is all you need.
\newblock {\em CoRR} abs/1706.03762.

\bibitem[\protect\citeauthoryear{Yu}{2017}]{Yu2017SummarizationWA}
Yu, H.
\newblock 2017.
\newblock Summarization with attention-based deep recurrent neural networks.

\end{thebibliography}

\end{document}